\documentclass[journal]{IEEEtran}
\usepackage{times}
\usepackage{amssymb}
\usepackage{helvet}
\usepackage{mathrsfs}
\usepackage{url}
\usepackage{courier}
\usepackage{booktabs,caption}
\usepackage{threeparttable}
\usepackage{amsmath}
\usepackage{float}
\usepackage{graphicx}
\usepackage{booktabs}
\usepackage{url}	
\usepackage{overpic}
\usepackage{color}
\usepackage[english]{babel}
\usepackage[T1]{fontenc}
\usepackage[utf8]{inputenc}
\usepackage{multirow}
\usepackage{diagbox}
\usepackage{authblk}
\usepackage[numbers, compress]{natbib}
\usepackage{times}
\usepackage{epsfig}
\usepackage{graphicx}
\usepackage{amsmath}
\usepackage{amssymb}
\usepackage{caption}
\usepackage{subcaption}
\usepackage{booktabs}
\usepackage{multirow}
\usepackage{graphics}
\usepackage{mathrsfs}
\usepackage{cuted}
\usepackage{graphbox}
\usepackage{capt-of}

\usepackage[pagebackref=true,breaklinks=true,colorlinks,bookmarks=false]{hyperref}

\begin{document}
\title{Back to the Roots: Reconstructing Large and Complex Cranial Defects using an Image-based Statistical Shape Model}

%

\author{Jianning Li, David G. Ellis, Antonio Pepe, Christina Gsaxner, Michele R. Aizenberg, Jens Kleesiek, \\ Jan Egger 

\thanks{We acknowledge CAMed (COMET K-Project 871132), FWF enFaced (KLI 678), FWF enFaced 2.0 (KLI 1044) and KITE (Plattform für KI-Translation Essen) from the REACT-EU initiative (\url{https://kite.ikim.nrw/}).}

\thanks{
J. Li, J. Kleesiek and J. Egger are with the Institute for Artificial Intelligence in Medicine (IKIM), Essen University Hospital, Girardetstraße 2, 45131 Essen, Germany.

J. Kleesiek and J. Egger are with the Cancer Research Center Cologne Essen (CCCE), University Medicine Essen, Hufelandstraße 55, 45147 Essen, Germany.

J. Kleesiek is with German Cancer Consortium (DKTK), Partner Site Essen, Hufelandstraße 55, 45147 Essen, Germany.

J. Li, A. Pepe, C. Gsaxner and J. Egger are with Computer Algorithms for Medicine Laboratory (Cafe), Graz, Austria.

J. Li, A. Pepe, C. Gsaxner and J. Egger are with the Institute of Computer Graphics and Vision (ICG), Graz University of Technology, Inffeldgasse 16c, 8010 Graz  Austria.

D. G. Ellis and M. R. Aizenberg are with the Department of Neurosurgery, University of Nebraska Medical Center, Omaha, NE, 68198, USA.

E-mails: Jianning.Li@uk-essen.de; Jan.Egger@uk-essen.de. Corresponding authors: Jianning Li and Jan Egger}}


\maketitle
\stripsep=5pt
\begin{strip}\centering\vspace{-4.5cm}
\includegraphics[width=\textwidth, vshift=0.2cm]{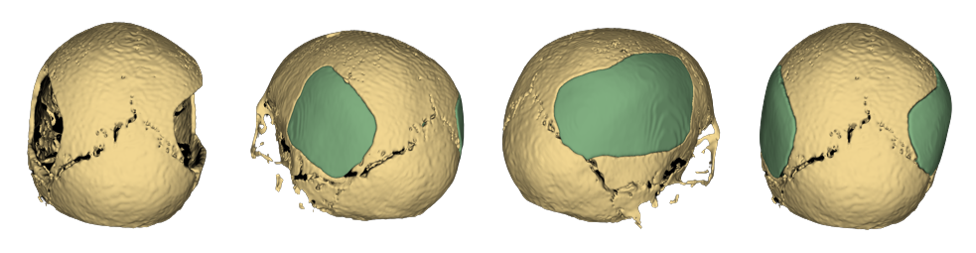}
\captionof{figure}{Example reconstruction of a skull with two defects. The skulls are shown in yellow and the implants in green.
\label{fig:teaser_new}}
\end{strip}

\textit{\textbf{Abstract-- }}\textbf{\textit{Designing implants for large and complex cranial defects is a challenging task, even for professional designers. Current efforts on automating the design process focused mainly on convolutional neural networks (CNN), which have produced state-of-the-art results on reconstructing synthetic defects. However, existing CNN-based methods have been difficult to translate to clinical practice in cranioplasty, as their performance on complex and irregular cranial defects remains unsatisfactory. In this paper, a statistical shape model (SSM) built directly on the segmentation masks of the skulls is presented. We evaluate the SSM on several cranial implant design tasks, and the results show that, while the SSM performs suboptimally on synthetic defects compared to CNN-based approaches, it is capable of reconstructing large and complex defects with only minor manual corrections. The quality of the resulting implants is examined and assured by experienced neurosurgeons. In contrast, CNN-based approaches, even with massive data augmentation, fail or produce less-than-satisfactory implants for these cases. Codes are publicly available at \url{https://github.com/Jianningli/ssm}}}.

\begin{IEEEkeywords}
Statistical shape model, Deep learning, Domain shift, Generalization, Cranial implant design, Cranioplasty, Craniotomy, Craniectomy
\end{IEEEkeywords}
\IEEEpeerreviewmaketitle

\section{Introduction}
Before deep learning gained wide popularity \citep{egger2021deep}, statistical shape model (SSM) and its variants (e.g., active shape models (ASMs) \citep{cootes1992active}, active blobs \citep{sclaroff1998active} and active appearance models (AAMs) \citep{cootes2001active}) were broadly adopted in medical reconstruction and segmentation tasks, such as the reconstruction of craniofacial defects \cite{SemperHogg2017VirtualRO, Fuessinger2019VirtualRO,Fuessinger2017PlanningOS,Lamecker2008VariationalAS,pimentel2020automated,ZhangKun2014} and human rib cage \cite{dworzak20103d}, and the segmentation of hip joints \cite{kainmueller2009articulated} and organs \cite{lamecker2002statistical}. In contrast to the latent shape features learned by deep neural nets, which are difficult to interpret, SSM offers the option to express a shape in an explicit manner, by linearly combining the mean shape and the principal modes of shape variations of a given shape pool. Surface meshes were common choices for anatomical shape representation in many SSM-based studies. Establishing dense point correspondence among the meshes is deemed the most demanding part in building an SSM, especially when the medical images, from which the meshes are derived, are of high resolution  \cite{heimann2009statistical}. Methods that establish points correspondence automatically are typically based on a mesh-to-mesh registration procedure (e.g., Iterative Closest Point \cite{besl1992method}), where the meshes are registered to a reference mesh through a similarity transformation (scaling, rotation and translation). However, popular state of the art segmentation methods and various medical applications are image-based \cite{ronneberger2015u,zhou2018unet++,li2018h,milletari2016v,pepe2020detection}. It is therefore desirable to circumvent the image-to-mesh conversion procedure and build an SSM directly on images \cite{reyneke2018construction,grauman2003inferring,bharath2018radiologic}. 

Automatic cranial implant design is another typical application that uses images as the initial shape representation \cite{li2021synthetic}. Existing deep learning-based methods usually train a deep neural net on hundreds of skull images with either synthetic defects  \cite{morais2019automated,li2021automatic,li2020baseline,li2021autoimplant} or clinical defects \cite{kodym2021deep}, depending on the availability of the clinical images. These approaches are data- and computation- intensive, and most importantly, the quality of their reconstructions for large and complex defects, which are common in cranioplasty, remains inadequate for clinical use \cite{ellis2021qualitative,mahdi2021u}. The failure can largely be attributed to domain shift: the synthetic defects in the training set have different distribution to that of the test set. Augmenting the training set intensively is a potentially practical and effective solution to the problem \cite{wodzinski2021improving}. However, current development in data augmentation-enhanced deep learning is still evaluated as substandard by clinical experts \cite{ellis2021qualitative,wodzinski2021improving}. 

A method that is independent from and insensitive to the defects may optimally avoid the domain shift problem in cranial defect reconstruction. To this end, we propose an SSM-based method for automatic cranial implant design. Unlike previous mesh-based SSM for craniofacial defect reconstruction \cite{Fuessinger2019VirtualRO,Fuessinger2017PlanningOS,pimentel2020automated}, our SSM is built directly on the volumetric skull images. We show that dense point correspondence among the training skull images can simply be achieved through an image registration and warping step, and the mean shape and shape variations of the skulls can be calculated thereafter. Besides the expected robustness against large and complex cranial defects, another favorable property of an SSM-based method is that the skull shapes can be expressed mathematically and explicitly, unlike deep learning-based approaches that try to learn an implicit and latent shape representation.  

The proposed SSM is evaluated on both synthetic defects and irregularly shaped clinical defects from three cranial implant design tasks. We show that even though deep learning still beats SSM on synthetic defect reconstruction, the performance of SSM is superior and stable when it comes to large and complex clinical defects.

\begin{figure*}[t]
\centering
\includegraphics[width=0.9\linewidth]{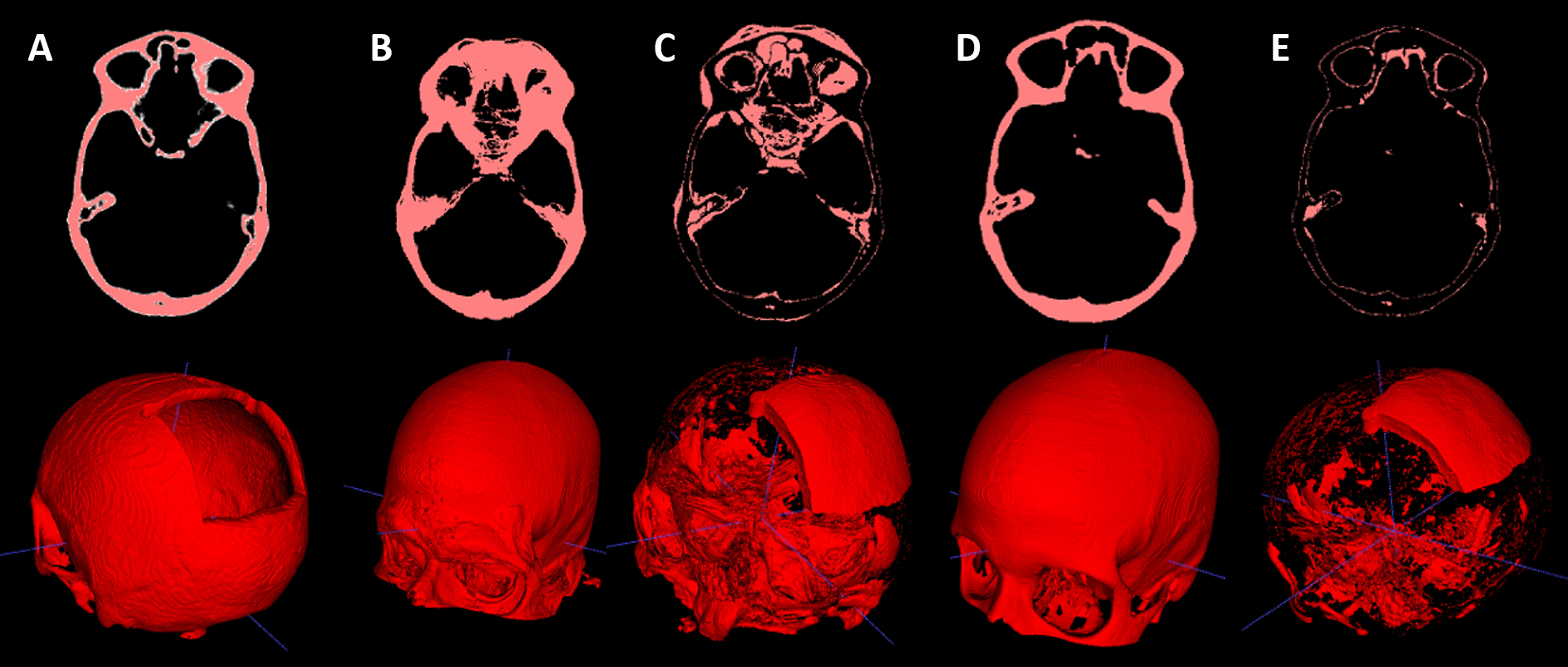}
\caption{A: the input defective skull. B: the mean skull ($\bar{S}$ (30)). C: the subtraction between B and A. D: the reference skull $x_j$. E: the subtraction between D and A.}
\label{fig:shape_warping}
\end{figure*}

\section{Method}
\subsection{Overview}
The main workload of building an SSM lies in finding the mean shape $\bar{S}$ and the primary shape variations $\mathbf{\phi}$ of a given shape pool, as specified by Equation~\ref{eq:ssm}:

\begin{equation}\label{eq:ssm}
S=\bar{S}+\sum_{i=1}^{C}\lambda_i\mathbf{\phi}_i
\end{equation}

$\lambda_i$ is the weight of variation $\mathbf{\phi}_i$. $C$ is the number of shape variations chosen for reconstructing the new shape $S$.  
Let $\mathbf{X}$ be a shape pool containing $C$ binary volumetric images $\mathbf{x}_i \in \{0,1\}^{D_i}$, in which the $j_{th}$ image $\mathbf{x}_j$ ($0<i, j\leqslant C$) is selected as a reference. $D_i$ is the image dimension. The non-zero voxels in these images constitute the geometry of the shapes and are regarded by default as the shape \textit{landmarks}. Therefore, establishing the point correspondences between all the images in the shape pool can be achieved by simply registering $\mathbf{x}_i$ ($i\neq j$) to the reference image $\mathbf{x}_j$:

\begin{equation}\label{eq:reg}
\mathbf{Tr}: x_i\rightarrow x_j
\end{equation}

$\mathbf{Tr}$ is a  transformation  that warps the images in $\mathbf{X}$ into the space of $\mathbf{x}_j$: $x'_i=\mathbf{T} (x_i)$, $x'_i \in R^{D_j}$. Let $X'=\begin{Bmatrix} x'_i \,| \, i\in Z, 0< i\leqslant C \end{Bmatrix}$, $\mathbf{X'} \in R^{ D_j \times C}$ be the set of warped shapes. The mean shape $\bar{S} \in R^{D_j}$ of $\mathbf{X}'$ is calculated as:

\begin{equation}\label{eq:reg}
\bar{S}=\frac{1}{C}\sum_{i}^{C}x'_i
\end{equation}

To extract the shape variations  $\mathbf{\phi}_i \in R^{D_j}$, principal component analysis (PCA) \citep{peason1901lines} is used. Let $\mathbf{\Phi}=\begin{Bmatrix}\phi_i\,|\,i\in Z, 0< i\leqslant C\end{Bmatrix}$,  $\mathbf{\Phi} \in R^{C \times D_j}$ ($C\ll D_j$) be the set of chosen shape variations. Transforming $X'$ into the PCA space can be achieved via:

\begin{equation}\label{eq:pca}
\mathbf{\Phi}\cdot X'=X_{pca}'
\end{equation}
$X_{pca}' \in R^{C \times C}$. The $X_{pca}'$ is given by the PCA function from the \textit{scikit-learn} package and $X'^{-1}$ is computed from the training set $X$. Therefore, we can calculate the variation matrix $\mathbf{\Phi}$ as follows\footnote{Note that we compute $\mathbf{\Phi}$ explicitly based on Equation~\ref{eq:eigenv}}:

\begin{equation}\label{eq:eigenv}
\mathbf{\Phi} =X_{pca}'\cdot X'^{-1}
\end{equation}

$X'^{-1}$ is a pseudo inverse of $X'$. Given a test shape $y$, it is first registered to the reference image $y'=\mathbf{Tr}(y), y' \in R^{D_j}$ and then mapped into the PCA space defined by the shape pool $X$: 

\begin{equation}\label{eq:weights}
\lambda =\mathbf{\Phi} \cdot y'
\end{equation}

$\lambda =\begin{Bmatrix}\lambda_i \,|\,i\in Z, 0< i\leqslant C\end{Bmatrix}$. We rescale $\lambda_i$ to [0,1] via: $(\lambda_i -min(\lambda))/max(\lambda)-min(\lambda)$. Given $\lambda$, $\mathbf{\Phi}$ and $\bar{S}$, the new shape can be computed according to Equation~\ref{eq:ssm}. In our work, $\mathbf{Tr}$ is chosen to be a similarity transformation. The reconstructed shapes can be warped to their original space via an inverse transformation $\mathbf{Tr}^{-1}$.

\subsection{Volumetric Shape Completion}
\subsubsection{Shape Warping}
An intuitive way to complete a defective shape $y$ is to warp it to the space of a complete shape $x_j$, which can be achieved through a registration process (i.e., $y'=\mathbf{Tr}(y)$). Since the anatomical landmarks of the two shapes are aligned because of registration, a following subtraction operation \footnote{Between two volumes of the same dimension.} between the two shapes can yield the missing portion of the defective shape $y_m$:

\begin{equation}\label{eq:implant}
y_m = x_j - y'
\end{equation}

The addition of $y'$ and $y_m$ produces the complete shape corresponding to $y'$. By inverting the registration, we can obtain the complete shape $y_c$ corresponding to $y$ in its original space: 

\begin{equation}\label{eq:inverse_reg}
y_c=\mathbf{Tr}^{-1}(y_m + y')
\end{equation}

The concept is similar to that of a template-based shape completion approach \cite{kraevoy2005template}, in which the missing part of a defective shape is taken from a complete template shape. The choice of the template shape affects the authenticity of $y_m$. Optimally, a shape $x_j$ that is general and representative of the shape class should be chosen as the template, to ensure that the registration error between $x_j$ and $y'$ is small and the missing part is taken from anatomically close regions on the template. The template shape can be from a single image like $x_j$, or the mean shape $\bar{S}$ of a shape pool, as specified in Equation~\ref{eq:reg}.

\subsubsection{SSM for Volumetric Shape Completion}
If the shape pool $X$ consists of complete shapes, while the test shape $y$ refers to a defective shape, applying Equation~\ref{eq:ssm} - Equation~\ref{eq:weights} would give the complete counterpart corresponding to $y$. In this sense, SSM can be used for shape completion tasks. In \cite{yu2021pca}, the authors used PCA for skull shape completion and 
showed that, by applying PCA and an inverse PCA consecutively to a defective skull, a complete skull can be obtained.  Equation~\ref{eq:ssm} and  Equation~\ref{eq:weights} give the mathematical explanation: the PCA computes the skull shape variations $\mathbf{\Phi}$ from the training samples and the weights $\lambda$ from the warped defective skull $y'$, while the inverse PCA, according to the the implementation of \textit{inverse\_transform} from the \textit{scikit-learn} package, computes:
\begin{equation}\label{eq:ssm_new}
S=\bar{S} +\lambda \cdot  \mathbf{\Phi}
\end{equation}
which is equivalent to Equation~\ref{eq:ssm}. Incorporating Equation~\ref{eq:weights} into Equation~\ref{eq:ssm_new} we get:

\begin{equation}\label{eq:ssm_new_new}
S=\bar{S} + \mathbf{\Phi}\cdot y'\cdot \mathbf{\Phi}=\bar{S}+y'\cdot\mathbf{\Phi}^T\cdot\mathbf{\Phi}
\end{equation}

In both \cite{yu2021pca} and Equation~\ref{eq:ssm}, the principal components of a defective skull are used as the weights of the shape variations. An obvious shortcoming is that, if the defects are too large, the principal components computed from a defective skull might not reflect the true distribution of the shape variations of a complete skull. For example, given a defective skull whose facial bone far outweighs the cranium due to a large cranial defect, the weight $\lambda_{k}$ of the variation concerning the facial area $\phi_{k}$ ($0<k\leqslant C$) would overwhelm the other variations, resulting in an inappropriate reconstruction of the region of interest (ROI, i.e., the cranium) for the cranial implant design task. To address this problem, we first use a shape completion network (denoted as DL)\footnote{The shape completion network refers to $N_1$ in \cite{li2020baseline, li2021learning}} to generate the complete counterpart of the defective skull $y'$ and use the completed skull to calculate $\lambda$:

\begin{equation}\label{eq:ssm_dl}
\lambda =\mathbf{\Phi} \cdot DL(y')
\end{equation}

However, whether using Equation \ref{eq:ssm_dl} for weight calculation has an positive effect on the results than using Equation \ref{eq:weights} depends on the quality of the DL reconstructions.

\section{Experiment and Results}

\subsection{Dataset and Metrics}
We evaluated our method on three datasets: the 11 clinical cases of defective skulls from Tasks 2 of the AutoImplant II challenge\footnote{\url{https://autoimplant2021.grand-challenge.org/}}, the 29 craniotomy skulls from MUG500+ \cite{li2021mug500+}, and the 110 test skulls with synthetic defects from Task 3 of AutoImplant II. To conform to the evaluation scheme of the AutoImplant II challenge, we measure the reconstruction accuracy using dice similarity coefficient (DSC), border DSC and 95 percentile Hausdorff distance (HD95).

The complete skulls from the training set of Task 3 were used as the shape pool $X$. The image dimension is $D_i= 512\times 512 \times Z_i$ \footnote{Except the last two cases from Task 2.} ($Z_i$ differs for different images). As calculating $\mathbf{\Phi}$ (Equation~\ref{eq:pca} and \ref{eq:eigenv}) from high resolution images is a computationally expensive process, we only used $C=30$ (out of 100) complete skulls for experiments involving $\mathbf{\Phi}$. The reference skull $x_j$ is chosen to be case \textit{001.nrrd} in the Task 3 training set and $Z_j=222$. All the training and test samples are registered to \textit{001.nrrd} through a similarity transformation $\mathbf{Tr}$.

\subsection{Reconstruction of Synthetic Cranial Defects}

\begin{figure}
\centering
\includegraphics[width=\linewidth]{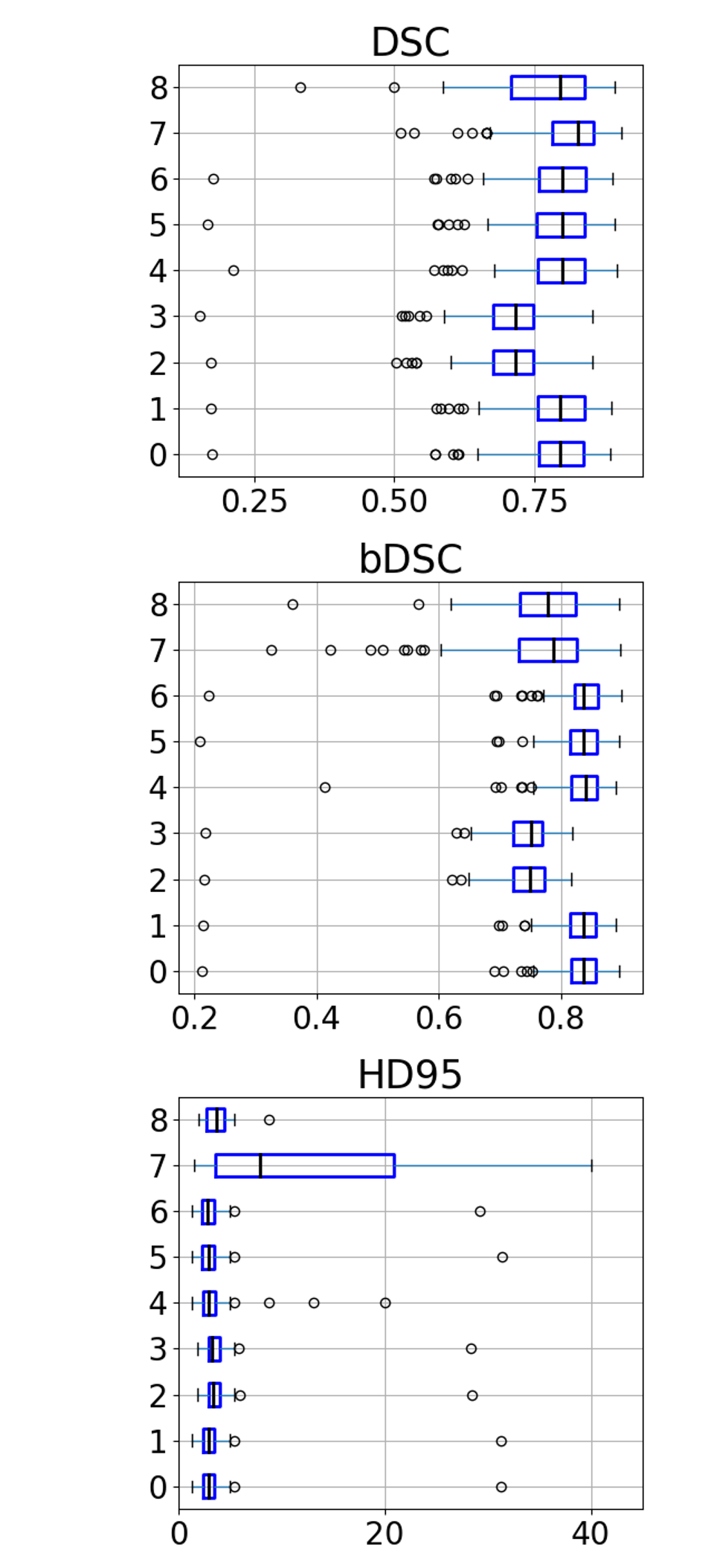}
\caption{Boxplots of DSC, bDSC and HD95. \textbf{0}: SSM (30), \textbf{1}: SSM (30) + DL, \textbf{2}: $\sum_{i=1}^{d_0}\lambda_i\Phi_i$  ($\lambda_i=1$), \textbf{3}: $\sum_{i=1}^{d_0}\lambda_i\Phi_i$, \textbf{4}: $\bar{S}$ (50), \textbf{5}: $\bar{S}$ (30), \textbf{6}: $x_j$, \textbf{7}: DL, \textbf{8}: \cite{yu2021pca}. }
\label{fig:boxplots}
\end{figure}

The 110 test skulls from Task 3 contain synthetic defects. In this experiment, we evaluate different methods of creating a skull template for shape warping: averaging 30 complete skulls ($\bar{S}$ (30)), averaging 50 complete skulls ($\bar{S}$ (50)) and using only a single skull ($x_j$). The 30 skulls are a subset of the 50 skulls. We also evaluate how the weights of shape variations $\lambda$ affect the reconstruction accuracy of SSM (Equation~\ref{eq:ssm}): $\lambda$ computed via Equation~\ref{eq:weights} (SSM (30)) and $\lambda$ computed via Equation~\ref{eq:ssm_dl} (SSM (30) + DL). The CNN-based shape completion network - DL is taken from \cite{li2021learning, li2020baseline} \footnote{\cite{li2021learning, li2020baseline} did not report bDSC and HD95. We calculate the two metrics in this paper based on the prediction files (implants) from \cite{li2021learning, li2020baseline}.}. Besides, shape reconstruction using only the shape variations is also evaluated:  $\sum_{i=1}^{d_0}\lambda_i\Phi_i$  ($\lambda_i=1$) and $\sum_{i=1}^{d_0}\lambda_i\Phi_i$. For the former, the $\lambda_i$ is set to one. For the latter, $\lambda_i$ is calculated regularly based on  Equation~\ref{eq:weights}. The DSC, bDSC and HD95 for these shape completion methods are reported in Table ~\ref{table:ablation} and Figure ~\ref{fig:boxplots}.

\begin{table}[h!]
\caption{Quantitative results (mean DSC, bDSC and HD95) on the 110 test cases of Task 3.}
\begin{center}
\scalebox{1.0}{
\begin{tabular}{ c|c|c|c} 
\toprule
Methods $\setminus$  Scores & DSC  & bDSC & HD95 (mm) \\
\midrule
$\bar{S}$ (30) &0.7840 &0.8265 &3.1989  \\ 
$\bar{S}$ (50) &0.7853 &0.8287 & 3.2447 \\ 
$x_j$  &0.7854 &0.8285 & 3.1700 \\ 
\hline
SSM (30)&0.7832&0.8255  & 3.2157 \\ 
SSM (30) + DL&0.7830 & 0.8253 & 3.2170 \\ 
DL \cite{li2021learning, li2020baseline} &0.8058&0.7638&13.2891\\ 
\hline
$\sum_{i=1}^{d_0}\lambda_i\Phi_i$  ($\lambda_i=1$)&0.7054 &0.7403 &3.6783 \\ 
$\sum_{i=1}^{d_0}\lambda_i\Phi_i$ &0.7064 &0.7411&3.6601  \\ 
\hline
L. Yu. et. al. \cite{yu2021pca} &0.7728&0.7716&3.6803\\
\hline
D. G. Ellis, et.al. \cite{ellis2020deep} &0.9440&-&-\\
M. Wodzinski et. al. \cite{wodzinski2021improving} &0.9329&0.9530&1.4781\\
\bottomrule
\end{tabular}}
\label{table:ablation}
\end{center}
\end{table}

\begin{figure}[t]
\centering
\includegraphics[width=0.9\linewidth]{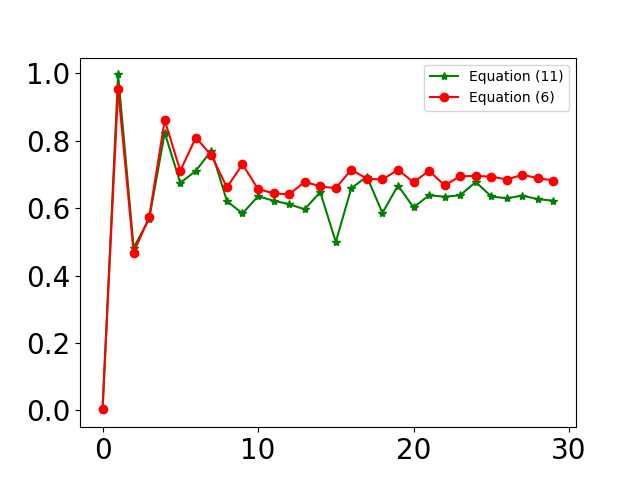}
\caption{$\lambda$ ($y$ axis)  computed based on Equation~\ref{eq:weights} and Equation~\ref{eq:ssm_dl} for different shape variations ($x$ axis refers to the index of variation). The averaged weights of the 110 test cases: $\frac{1}{110}\sum_{k=1}^{110}\lambda_{i,k}$ is used for this plot.}
\label{fig:weights}
\end{figure}

The results bear important implications: \textbf{(1)} Using a single skull image $x_j$ as the template for shape warping can produce reasonable results, qualitatively and quantitatively (third row, Table ~\ref{table:ablation}, and Figure ~\ref{fig:shape_warping} E). However, it should be noted that the conclusion applies only to the cranial implant design task, where the ROI to reconstruct is the structurally uncomplicated cranium. The single image-based template would likely to fail for reconstructing individual-specific facial bones, which contain complex and subtle structures. \textbf{(2)} Shape template derived from a single shape ($x_j$) produces comparable results to that from a mean shape averaged from 30 ($\bar{S}$ (30)) or 50 ($\bar{S}$ (50)) images. Figure ~\ref{fig:shape_warping} gives an example of the results obtained using shape warping. We can see that $\bar{S} (30)$ (Figure ~\ref{fig:shape_warping} B) shows no noticeable difference on the cranium compared to $x_j$ (Figure ~\ref{fig:shape_warping} D). As a result, subtracting the input from the templates (Equation~\ref{eq:implant}) produces similar implants. The main difference lies in the facial area and the interior subtle structures (Figure ~\ref{fig:shape_warping} C and E). Applicable to both statements (1) and (2), the reconstruction accuracy depends largely on how well the target matches with the template on the ROI (e.g., cranium or facial bones) during the warping and registration process. It is relatively easier to register among different craniums than different facial bones. For a facial reconstruction task (e.g., facial implant design), a mean shape, as illustrated in Figure ~\ref{fig:shape_warping} (B), possesses significantly more facial landmarks than a single image, which might potentially make the facial registration more accurate. \textbf{(3)} The weights of the shape variations affect the accuracy of cranium reconstruction. An analysis of the shape variation matrix $\Phi$ reveals that, around 53\% of the $C=30$ variations are related to the full skull and the remaining are associated with the facial area only, which do not contribute to the cranium reconstruction. The weight distribution (calculated based on Equation~\ref{eq:weights}) of $\sum_{i=1}^{d_0}\lambda_i\Phi_i$ is shown in Figure ~\ref{fig:weights}. We can see that the largest three weights in $\sum_{i=1}^{d_0}\lambda_i\Phi_i$ correspond to the $i=2,5,7$ variations: $\mathbf{\phi}_2$ (facial bones), $\mathbf{\phi}_5$ (full skull) and $\mathbf{\phi}_7$ (full skull). The results presented in the $\sum_{i=1}^{d_0}\lambda_i\Phi_i$  ($\lambda_i=1$) and $\sum_{i=1}^{d_0}\lambda_i\Phi_i$ rows of Table ~\ref{table:ablation} only show marginal differences, meaning that the two full skull-related variations ($\mathbf{\phi}_5$ and $\mathbf{\phi}_7$) already carries the information necessary for reconstructing a complete skull. Figure ~\ref{fig:weights} also shows the weight distribution $\lambda$ calculated based on Equation~\ref{eq:ssm_dl}. The variations corresponding to the three largest weights are $\mathbf{\phi}_2$ (facial bones), $\mathbf{\phi}_5$ (full skull) and $\mathbf{\phi}_8$ (full skull). Compared to 'SSM (30)', the deviation in weight distribution for 'SSM (30) + DL' leads to marginal decline in the quantitative scores, meaning that a better network or a training method should be chosen in order that the DL can benefit 'SSM (30)'. \textbf{(4)} In comparison to the deep learning-based approaches \cite{ellis2020deep,wodzinski2021improving}, the shape warping- and SSM-based methods achieve inferior results on synthetic defects quantitatively. However, it should be noted that both \cite{ellis2020deep} and \cite{wodzinski2021improving} used an intensively augmented dataset during training, while only 30 images were used to build the SSM. 

\begin{figure}[ht]
\centering
\includegraphics[width=0.9\linewidth]{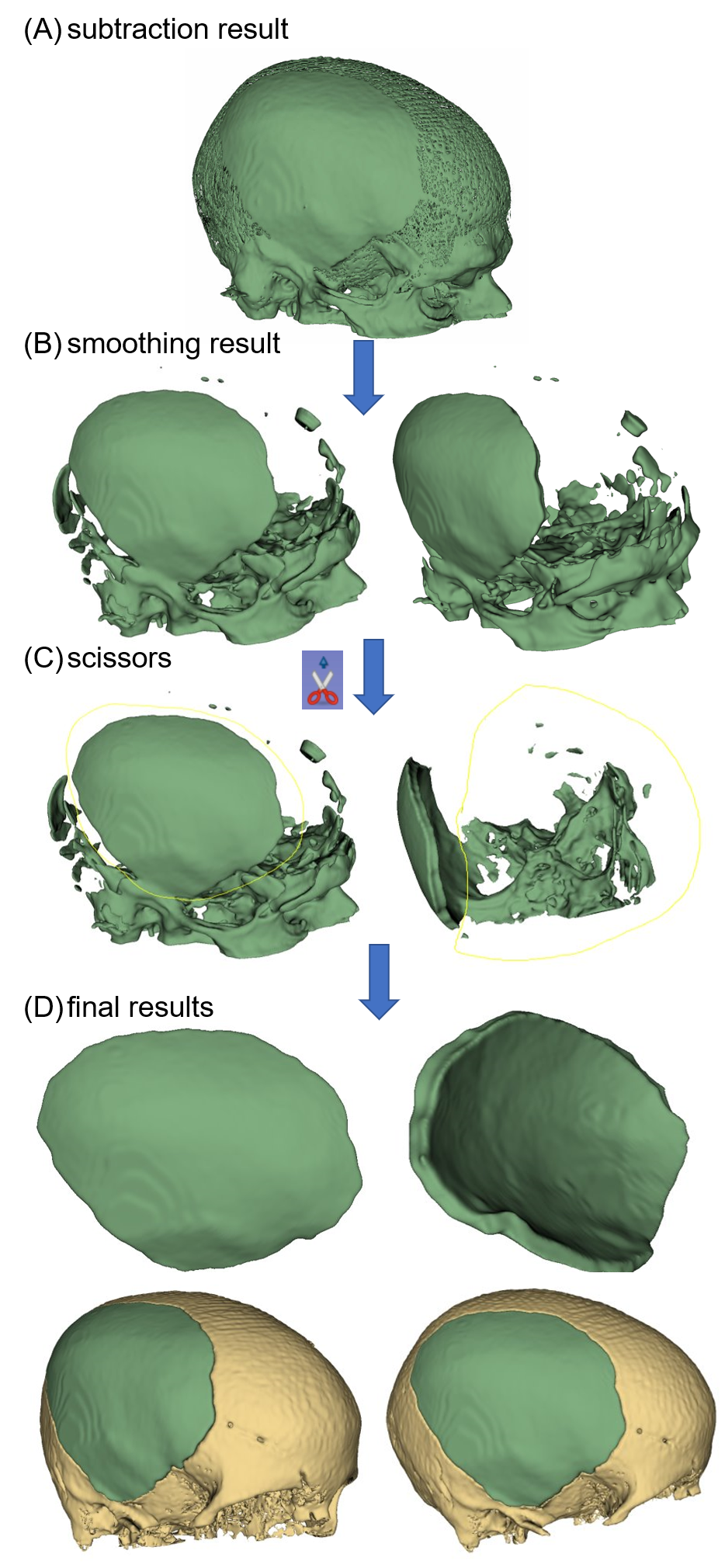}
\caption{Manually extract the implants from the subtraction results using 3D Slicer.}
\label{fig:post_processing}
\end{figure}

\begin{figure*}
\centering
\includegraphics[width=0.9\linewidth]{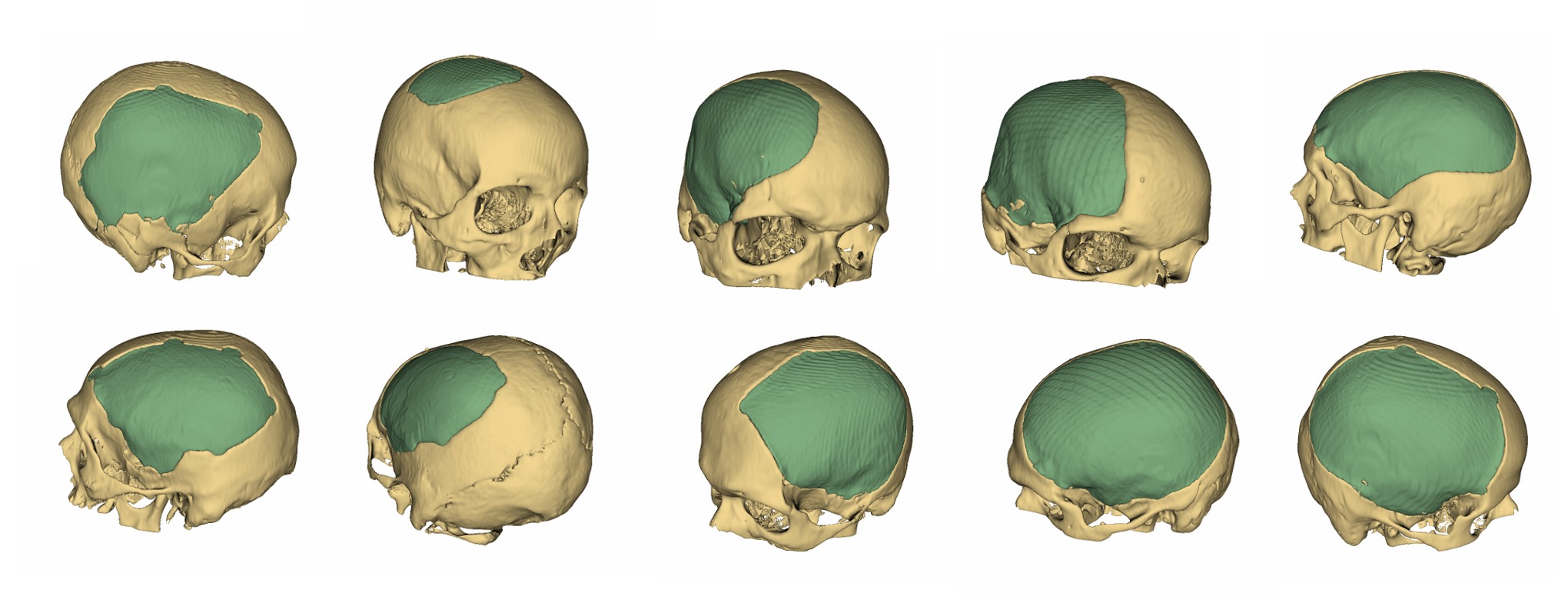}
\caption{Exemplary results (from $\bar{S}$ (50)) on the craniotomy skulls from the MUG500+ dataset. The 29 generated implants can be downloaded at \url{https://doi.org/10.6084/m9.figshare.19328816.v3}}.
\label{fig:mug500_results}
\end{figure*}

\begin{figure*}
\centering
\includegraphics[width=0.85\linewidth]{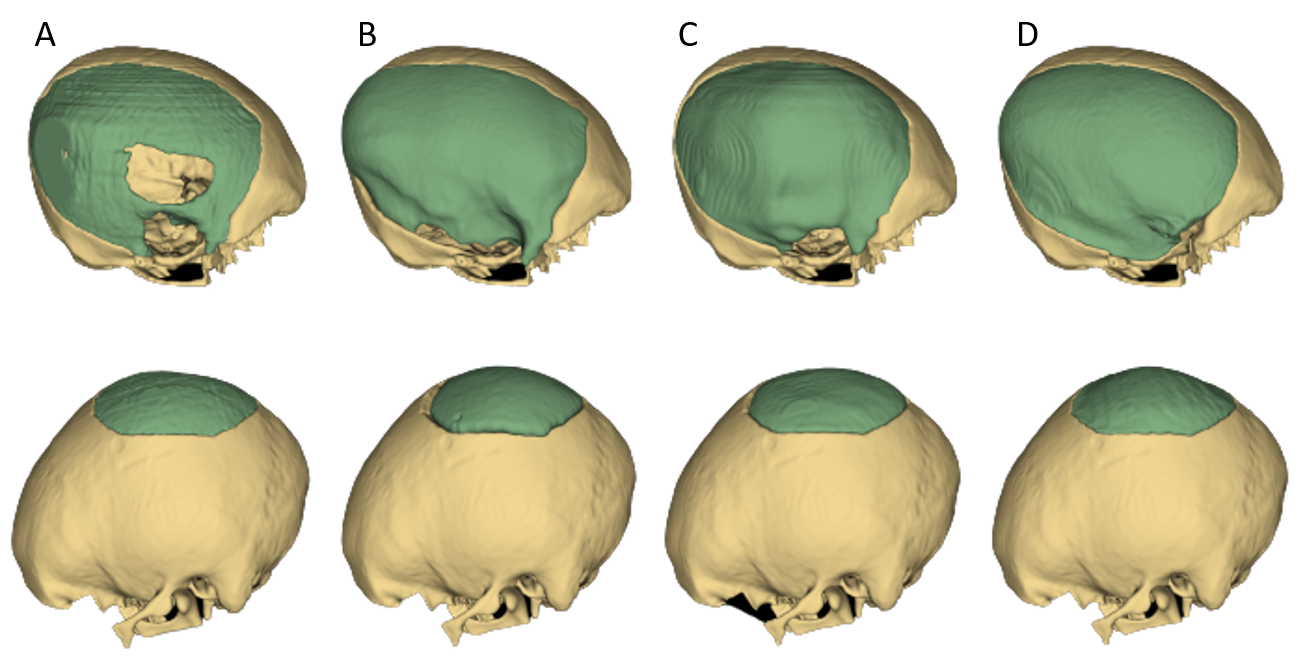}
\caption{Qualitative comparison between different implant design methods on Task 2@AutoImplant II. (A) \cite{wodzinski2021improving} (B) \cite{yu2021pca} (C) \cite{mahdi2021u} (D) Ours. The 11 generated implants can be downloaded at \url{https://doi.org/10.6084/m9.figshare.19328816.v3}}.
\label{fig:task2_results}
\end{figure*}

\subsection{MUG500+}
This section presents the implant generation results on the craniotomy skulls from the MUG500+ dataset \cite{li2021mug500+}. Figure ~\ref{fig:post_processing} shows how to manually post-process an implant (Figure ~\ref{fig:post_processing} (A)) generated by subtracting the input defective skull from the skull reconstructed by SSM (30). First, a median smoothing filter is applied to the subtraction result to (partly) disconnect the implant from the noise (Figure ~\ref{fig:post_processing} B). The smoothing kernel size should be chosen individually. Second, the \textit{scissors} functionality is used to erase the delineated area (Figure ~\ref{fig:post_processing} C) to fully remove the noise and extract the implant. Figure ~\ref{fig:post_processing} (D) shows the post-processing result. Step (B)  and (C) is done manually using 3D Slicer (\url{https://www.slicer.org/}) \cite{egger2013gbm}. Alternatively, the implant can be extracted automatically by applying morphological opening and connected component analysis consecutively to the subtraction result. However, it is recommended to follow the manual post-processing workflow in Figure ~\ref{fig:post_processing} for an optimal outcome \footnote{For example, in Figure ~\ref{fig:post_processing} (C), the implant is still connected to the facial bones after smoothing. Where to erase in the connecting region should optimally be determined according to the defect shape.}. Figure ~\ref{fig:mug500_results} presents the automatically generated implants for some large and complex defects in the MUG500+ dataset. The implants are generated by $\bar{S}$ (50) and manually post-processed. We can see that some of the defects are large enough to cover half of the cranium or have rather irregular shapes. Nonetheless, the defects are still satisfactorily reconstructed. Notably, Figure ~\ref{fig:teaser_new} (the teaser image) shows that the method is still effective when multiple large defects exit on the craniotomy skull. The completed skulls obtained according to Equation ~\ref{eq:inverse_reg} preserve the anatomical aesthetics of normal human skulls.  

The MUG500+ dataset contains the manually designed implants (i.e., surface models in \textit{.stl} format) for the 29 craniotomy skulls. The manual designs are converted to images (\textit{.nrrd}) for a quantitative comparison with our automatic designs. The results are given in Table ~\ref{table:mug500}. Keep in mind that, while interpreting the scores, the quantitative results are only a reflection of how well our automatic reconstructions match with the manual designs, which is subjective and experience-dependent\footnote{For example, for large defects, different designers might decide differently for the curvature of the implants based on their own aesthetic views.}. Experts' manual and qualitative evaluation of the implants has closer correlation with the true quality of the reconstructions \cite{ellis2021qualitative}.

\begin{table}[h!]
\caption{Quantitative results (produced by $\bar{S}$ (50)) for the MUG500+ craniotomy dataset.}
\begin{center}
\scalebox{1.0}{
\begin{tabular}{ c|c|c|c} 
\toprule
Methods $\setminus$  Scores & DSC  & bDSC & HD95 \\
\midrule
$\bar{S}$ (50)&0.5471& 0.5761 & 5.0000 \\ 

\bottomrule
\end{tabular}}
\label{table:mug500}
\end{center}
\end{table}

\subsection{Task2}
To show the utility of our methods on real clinical cases, the implant designs were created for the 11 clinical defective skulls from the Task 2 of the AutoImplant II challenge. As described in Ellis et al. \cite{ellis2021qualitative}, the implant designs were quantitatively compared to reconstructions from postoperative imaging of the actual implant the patients' received. Table ~\ref{table:task2} shows these quantitative results from $\bar{S}$ (50) and SSM (30), as well as from the AutoImplant II submissions. The $\bar{S}$ (50) and SSM (30) had the best Hausdorff 95 scores than all other submissions but scored worse than some other submissions in the dice similarity and boundary dice similarity scores.

\begin{table}[h!]
\caption{Quantitative results for Task 2 of the AutoImplant II challenge.}
\begin{center}
\scalebox{1.0}{
\begin{tabular}{ c|c|c|c} 
\toprule
Methods $\setminus$  Scores & DSC  & bDSC & HD95 \\
\midrule
$\bar{S}$ (50)&0.5007&0.4449  & 8.2539 \\ 
SSM (30) &0.5055 &0.4470  &7.9042  \\ 
M. Wodzinski. et al. \cite{wodzinski2021improving} & 0.5241&0.4823  & 54.5165 \\
L. Yu. et al. \cite{yu2021pca} &0.5118 &0.4547 &8.3486  \\
H. Mahdi. et al. \cite{mahdi2021u} &0.3028 & 0.3092 & 71.4193 \\

\bottomrule
\end{tabular}}
\label{table:task2}
\end{center}
\end{table}

However, comparing the implant designs to the reconstructions of the implants from the postoperative CT imaging do not necessarily serve as a reliable metric for the quality of the implant designs \cite{ellis2021qualitative}. For this reason, the implant designs were also qualitatively evaluated by experienced neurosurgeon, MRA. The implant designs were judged based on completeness, false positive area, fit, and overall feasibility as described in Ellis et al. \cite{ellis2021qualitative}. As shown in Table ~\ref{table:task2_qual}, the $\bar{S}$ (50) implant designs had better overall feasibility, better fit, and less false positive area than the submissions from the Autoimplant II challenge. While none of the submissions from the Autoimplant II challenge were deemed feasible without modifications, 4 out of 11 of the $\bar{S}$ (50) designs were deemed feasible with only minor flaws. Therefore, the $\bar{S}$ (50) designs represent a substantial improvement in the clinical feasibility of implant designs. The main issues plaguing the $\bar{S}$ (50) implant designs were that they did not always extend far enough in the superior direction to fully restore the natural skull shape and that the implants were often too thick.

\begin{table}[h!]
\caption{Qualitative evaluation scores for Task 2 of the AutoImplant II challenge by neurosurgeon MRA. The scores have been normalized such that 0 is the lowest possible score and 1 is the highest possible score. Completeness (Comp) evaluates the amount of the defect that the implant design covers. False positive area (FPA) evaluates the amount of amount of implant design outside of the defect area. Fit evaluates the shape of the implant design relative to the defect. Feasibility evaluates whether the implant design could be used in surgery. See Ellis et al. for the qualitative analysis methods \cite{ellis2021qualitative}.}
\begin{center}
\begin{tabular}{ c|c|c|c|c} 
\toprule
Methods $\setminus$  Scores & Comp & FPA & Fit & Feasibility \\
\midrule
$\bar{S}$ (50)&0.89& 0.73& 0.64& 0.62 \\ 
M. Wodzinski. et al. \cite{wodzinski2021improving}&0.93&0.57&0.55&0.42 \\
L. Yu. et al. \cite{yu2021pca} &0.80  & 0.59 & 0.36 & 0.42  \\
H. Mahdi. et al. \cite{mahdi2021u} &0.76 & 0.43 & 0.45 & 0.33 \\

\bottomrule
\end{tabular}
\label{table:task2_qual}
\end{center}
\end{table}

\section{Discussion}
In automatic cranial implant design, deep learning-based approaches that rely on a \textit{defect-complete} or \textit{defect-implant} pair for training often fail to generalize to large and complex cranial defects in the test set, since the synthetic defects used during training have different distributions to the real defects during evaluation (i.e., domain shift). One popular solution to this problem is resorting to intensive data augmentation: augment the defects \cite{li2021automatic, kodymskullbreak} and/or augment the skull images  \cite{wodzinski2021improving, ellis2020deep}. The former tries to create realistic synthetic defects for training. Data augmentation has shown to be effective to the generalization problem for deep learning in automatic cranial implant design. However, the computational cost is substantially increased. Besides, the patient-specific cranial defects can depict considerable variations among individuals, making it impractical to cover all defect patterns through augmentation. The SSM-based approach can circumvent the defect-related generalization problem, as an SSM relies only on the complete skulls available in the training set for training. Therefore, an SSM is insensitive to and unaffected by the changes in defect patterns in the test cases. One factor affecting the performance of an SSM is the registration accuracy. Since human craniums are structurally uncomplicated and topologically stable among individuals compared to the defects and facial bones, precise registration among different craniums is highly achievable. Therefore, an SSM or simply a shape warping-based approach, though methodologically simple, is ideal for the cranial defect reconstruction task. 

Furthermore, due to the black-box nature of deep learning, the shape features learned by a deep neural network are often entangled and lacking interpretability. In contrast, the basic components of an SSM - the mean shape and shape variations, are explicitly expressed and therefore highly comprehensible to humans. In clinical settings, interpretability and transparency adds to the trust in the methods.

\section{Conclusion}
As an alternative to mesh-based SSM, we demonstrate in this paper that an SSM can be built directly on (volumetric) skull images. The image-based SSM of the skull is evaluated on three cranial defect reconstruction tasks, and the results reveal that even if state of the art deep learning-based approaches still prevail in reconstructing synthetic defects, our SSM-based method shows advantages in the reconstruction of large and complex defects that are common in cranioplasty. Besides, the SSM-based methods are not dependent on large quantities of training data as deep learning, making the proposed approach highly scalable and applicable in a clinical setting.


\bibliographystyle{IEEEtran}
\bibliography{references}
\end{document}